\pdfoutput=1
\documentclass[11pt]{article}

\usepackage[final]{acl}
\usepackage{times}
\usepackage{latexsym}

\usepackage[T1]{fontenc}
\usepackage[utf8]{inputenc}

\usepackage{microtype}

\usepackage{inconsolata}

\usepackage{graphicx}
\usepackage{amssymb}
\usepackage{amsmath}
\usepackage{multirow}
\usepackage{booktabs}
\usepackage{colortbl}
\usepackage{makecell}
\usepackage{subfigure}

\usepackage{hhline}
\usepackage{color}
\usepackage{colortbl}
\usepackage{booktabs}
\usepackage{tabularx}
\usepackage{makecell}

\title{Exploring the Compositional Deficiency of Large Language Models in Mathematical Reasoning Through Trap Problems
}

\author{Jun Zhao$^{1}$\footnotemark[1]\;,\ \ Jingqi Tong$^{1}$\footnotemark[1]\;,\ \ Yurong Mou$^{1}$,\ \ Ming Zhang$^{1}$,\ \ \\\textbf{Qi Zhang}$^{1,2}$\footnotemark[2], \textbf{Xuanjing Huang}$^{1,2}$\\
  $^1$School of Computer Science, Fudan University\\
  $^2$Shanghai Key Laboratory of Intelligent Information Processing, Fudan University\\
  \texttt{\{zhaoj19,qz\}@fudan.edu.cn,jqtong23@m.fudan.edu.cn}}
  
\begin{document}
\maketitle
\renewcommand{\thefootnote}{\fnsymbol{footnote}}
\footnotetext[1]{Equal Contributions.}
\footnotetext[2]{Corresponding authors.}
\footnotetext[3]{https://github.com/tongjingqi/MathTrap}
\begin{abstract}
Human cognition exhibits systematic compositionality, the algebraic ability to generate infinite novel combinations from finite learned components, which is the key to understanding and reasoning about complex logic. In this work, we investigate the compositionality of large language models (LLMs) in mathematical reasoning. Specifically, we construct a new dataset \textsc{MathTrap}\footnotemark[3] by introducing carefully designed logical traps into the problem descriptions of MATH and GSM8K. Since problems with logical flaws are quite rare in the real world, these represent ``unseen'' cases to LLMs.
Solving these requires the models to systematically compose (1) the mathematical knowledge involved in the original problems with (2) knowledge related to the introduced traps. Our experiments show that while LLMs possess both components of requisite knowledge, they do not \textbf{spontaneously} combine them to handle these novel cases. We explore several methods to mitigate this deficiency, such as natural language prompts, few-shot demonstrations, and fine-tuning. Additionally, we test the recently released OpenAI o1 model and find that human-like `slow thinking' helps improve the compositionality of LLMs.
%We find that LLMs' performance can be improved through the above external intervention. 
Overall, systematic compositionality remains an open challenge for large language models.
%when explicitly alerted to the presence of traps, LLM performance improves significantly. Overall, systematic compositionality remains an open challenge for large language models.
%We find that LLMs (1) can accurately reason about the problems in Math and GSM8k; (2) recognize that the planted traps are logically inconsistent; (3) fail to realize the traps in the reasoning process. This suggests that much of the LLM performance on mathematical reasoning benchmarks arises from memorizing reasoning paths in the training data rather than a deep understanding and systematic compositionality of mathematical concepts. We also identify several approaches to mitigate this deficiency, such as (1) providing prompts that direct the model's attention to the traps in the problems as instructions, or (2) teaching the model the reasoning logic of problems containing traps through in-context learning and fine-tuning.
\end{abstract}
\section{Introduction}

\begin{table*}[h]
    {  \footnotesize \begin{tabularx}{\linewidth}{cc X} 
  
    \toprule 
    \multirow{6}{*}{\makecell[c]{\textbf{Concept} \\\textbf{Undefined}\\(16\%)}} &Original Problem  & {In right triangle $XYZ$ with $\angle YXZ = 90^\circ$, $XY = 24$ and $YZ = 25$. Find  \colorbox {yellow}{$\tan Y$}.}\\
    &Trap Problem  & {In right triangle $XYZ$ with $\angle YXZ = 90^\circ$,  $XY = 24$ and $YZ = 25$. Find  \colorbox {yellow}{$\tan X$}.}\\
    &Conceptual Problem& {
Does $\tan90^\circ$ exist?}\\
    \hline\hline
    \multirow{6}{*}{\makecell[c]{\textbf{Missing } \\\textbf{Condition}\\(6\%)}} &Original Problem  & {Natalia sold 48 clips in April and half as many clips in May. How many clips did Natalia sell altogether in April and  \colorbox {yellow}{May}?}\\
    &Trap Problem  & {Natalia sold 48 clips in April and half as many clips in May. How many clips did Natalia sell altogether in April and  \colorbox {yellow}{June}?}\\
    &Conceptual Problem& {Given the sales figures for May and June, can the sales figures for April and June be calculated?}\\
    \hline\hline
    \multirow{5}{*}{\makecell[c]{\textbf{Direct } \\\textbf{Contradiction}\\(24\%)}} &Original Problem  & {An equilateral triangle has a perimeter of 30 centimeters. Calculate its area.}\\
    &Trap Problem  & {An equilateral triangle has a perimeter of 30 centimeters \colorbox {yellow}{and a height of 10} \colorbox{yellow}{centimeters.} Calculate its area.}\\
    &Conceptual Problem& {Can the height of an equilateral triangle be equal to its side length?}\\
    \hline\hline
    \multirow{3}{*}{\makecell[c]{\textbf{Indirect} \\\textbf{ Contradiction}\\(38\%)}} &Original Problem  & {Find the solution of the equation $x^2+x=3$.}\\
    &Trap Problem  & {Find the \colorbox {yellow}{integer} solution of the equation $x^2+x=3$.}\\
    &Conceptual Problem& {Is the $\sqrt{13}$ an integer?}\\
    \hline\hline
    \multirow{5}{*}{\makecell[c]{\textbf{Violating} \\\textbf{ Common Sense}\\(15\%)}} &Original Problem  & {Max picks \colorbox {yellow}{2} different cards without replacement from a standard 52-card deck. What is the probability that the cards are of different suits?}\\
    &Trap Problem  & {Max picks \colorbox {yellow}{5} different cards without replacement from a standard 52-card deck. What is the probability that the cards are of different suits?}\\
    &Conceptual Problem& {Is it possible to pick five different suits of cards from a standard deck of playing cards?}\\
  \bottomrule 
 \end{tabularx} }
    \caption{Overview of the \textsc{MathTrap} Dataset. The first column represents the five introduced trap types and their percentages in the dataset. The \colorbox {yellow}{yellow highlighted} text emphasizes the difference in problem descriptions before and after introducing traps. Additionally, we annotate Conceptual Problems to test whether models possess trap-related knowledge. \textit{We hope that if a model can accurately answer both the Original Problems and the Conceptual Problems, it will also be able to accurately answer the Trap Problems.} Appendix section \ref{app_sec:category} provides definitions of the trap types, and Table \ref{app_tab:explation} offers explanations for these 5 example traps. We have included GPT-4-0125-preview's responses to selected problem from the table in Appendix Tables \ref{app_tab:output_of_ChatGPT4_trap_1}-\ref{app_tab:output_of_ChatGPT4_original_2}.}
    \label{tab:case}
\end{table*}
%而大模型却很容易在该类组合问题上出错
Humans excel at learning fundamental concepts and skills, systematically combining them to solve new problems. %For example, when someone understands that ``$\tan 90^\circ$ is undefined'' and that ``In right triangle $XYZ$ with $\angle YXZ = 90^\circ$,'' they recognize that ``$\tan X$ cannot be determined'', even without being explicitly told. 
For instance, when a person possesses (a) the knowledge of how to solve quadratic equations with one variable, and (b) the understanding of what integers are, they can combine these two domains of knowledge to tackle the problem ``Find the integer solutions of $x^2 + x = 3$.'' They would first solve the equation, and then determine whether the obtained solutions are integers or not.
\citet{fodor1988connectionism} had a famous viewpoint that artificial neural networks lack this compositionality, and thus cannot serve as reliable cognitive models. Current LLMs have achieved unprecedented success on tasks requiring complex reasoning \cite{guo2024deepseekcoder,toshniwal2024openmathinstruct1}. We wonder whether compositionality still poses a significant challenge for LLMs? \footnote{Due to space constraints, we provide a detailed description of existing research on LLMs' ability to learn and combine knowledge in Appendix \ref{app:limitations}: Related Work.}

Toward this goal, we construct a new \textsc{MathTrap} dataset by introducing carefully designed logical traps into the original problems of the MATH \cite{DBLP:journals/corr/abs-2103-03874} and GSM8K \cite{DBLP:journals/corr/abs-2110-14168} datasets. 
For example, by modifying the original problem ``Find the solution of the equation $x^2 + x = 3$'' to ``Find the \textbf{integer} solution of the equation $x^2 + x = 3$,'' the model needs to combine (a) the knowledge involved in the original problem (how to solve quadratic equations with one variable) and (b) the knowledge about the trap (the definition of integers) to handle these trap problems (in fact, the original equation has no integer solutions). Another reason for evaluating compositionality through trap problems is that these problems rarely appear in the real world, so it is unlikely that LLMs provide the correct answers solely by following the trained reasoning paths.

%Trap problems rarely appear in the real world, so it is unlikely that LLMs provide the correct answers solely by following the reasoning paths seen during training.
%Handling these trap problems requires models to systematically compose (a) the knowledge involved in the original problems (e.g., the definition of right triangle) and (b) the knowledge about the traps (e.g., $\tan90^\circ$ is not defined).
We conduct comprehensive tests on leading LLMs and recruit 43 undergraduate students from top universities as human controls. We find that LLMs and humans exhibit strikingly different behavioral patterns when dealing with trap problems. Despite possessing both (a) and (b) knowledge components, LLMs fail to \textbf{spontaneously} compose them to handle trap problems, while humans can. This suggests that tasks requiring compositional generalization remain challenging for current LLMs. Furthermore, the ability of well-aligned LLMs to handle trap problems can be elicited through external interventions, such as natural language prompts, few-shot demonstrations, and supervised fine-tuning. Furthermore, we find that the human-like `slow thinking' demonstrated by OpenAI's o1 \cite{o1} also helps improve the compositionality of LLMs. Nevertheless, systematic compositionality remains an open challenge for current LLMs.

The contributions of this work are threefold: (1) We investigate the compositional generalization of LLMs on mathematical reasoning, and demonstrate their stark under performance compared to humans. (2) An effective method to construct `unseen problems' by introducing traps into original problems, and a dataset called \textsc{MathTrap} that cannot be solved by simply following the reasoning paths seen during training. (3) Comprehensive experiments exploring the impact of model size, the degree of alignment, and external interventions on performance on the \textsc{MathTrap}.
%and a dataset called MathTrap that truly reflects LLMs' compositional abilities. 

%The contributions of this paper are as follows: (1) we investigate the compositional generalization of LLMs and show that compositional systematicity remains an open challenge; (2) we carefully design 5 types of traps and construct the \textsc{MathTrap} dataset for evaluating the compositional abilities of LLMs; (3) comprehensive experiments explore how model scales, and compositional external interventions affect the performance on \textsc{MathTrap}.
\section{Background and Definition}
%Compositionality is a complex concept spanning multiple domains. 
In this section, we provide the definition of compositionality discussed in this paper, based on Hilbert's formal deductive systems \cite{hilbert}:

\noindent\textbf{Definition 1}. (\textit{Hilbert's Formal Deductive System}) This system consists of (1) a syntax $\mathcal{G}$, specifying which derivation statements are legal, (2) a set of inference rules $\mathcal{R}$ \footnote{We refer to the axioms and inference rules as knowledge.}, clearly stating how new facts (or theorems) can be derived from existing facts (axioms or already proved theorems), and (3) axioms: a predetermined set $\mathcal{A}$ of established facts.

Under this deductive system, reasoning is defined as the process of deriving new facts from existing facts and rules.

\noindent\textbf{Definition 2}. (\textit{Compositionality in Mathematical Reasoning})
Suppose problem sets $Q_1$, $Q_2$, $Q_3$ are described using the same syntax $\mathcal{G}$, and $Q_1$ and $Q_2$ can derive final answers through tuple ($\mathcal{R}_1$, $\mathcal{A}_1$), ($\mathcal{R}_2$, $\mathcal{A}_2$) respectively, while $Q_3$ requires reasoning using $\mathcal{R}_1\cup \mathcal{R}_2$ based on $\mathcal{A}_1\cup \mathcal{A}_2$ (or a subset) to derive the final answer. If a reasoning engine can solve $Q_1$ and $Q_2$, we say it possesses compositionality if it can solve $Q_3$.

\section{The \textsc{MathTrap} Dataset}

Existing datasets for evaluating compositionality are limited to symbolic reasoning with semantic decoupling \cite{article,dziri2023faith}. However, the semantics of language play a crucial role in the reasoning process of LLMs \cite{tang2024on}. Our \textsc{MathTrap} dataset aims to evaluate the compositionality of LLMs on semantically rich math word problems. More importantly, the `unseen' feature of our dataset prevents models from simply following the trained reasoning paths to arrive at solutions.

\subsection{Dataset Composition}
\label{app_sec:category}
\noindent\textbf{Sample Composition}: 
As illustrated in Table \ref{tab:case}, each sample in \textsc{MathTrap} can be viewed as a problem triplet:

\noindent\textit{(1) Original problem}: Sampled from the MATH and GSM8K datasets. These problems are used to evaluate the model's grasp of math knowledge from these datasets.

\noindent\textit{(2) Concept Problem}: Manually crafted to assess the model's understanding of the trap concepts to be introduced. These problems are intentionally simple, requiring only knowledge of the trap concept to solve.

\noindent \textit{(3) Trap Problem}: Created by manually introducing logical traps into the original problems. These are designed to evaluate the model's compositional  generalization ability. Solving these problems requires the model to systematically combine knowledge from the original math problem with the introduced trap concept.

The \textsc{MathTrap} dataset consists of two subsets: \textit{Public} and \textit{Private}. The \textit{Public} subset contains $105$ problem triplets. Using GPT-4, we paraphrase these samples to expand the dataset to $1,000$ problem triplets, which are used in all fine-tuning experiments discussed in this paper. We manually verify the quality of the subset. Please refer to Table \ref{app_tab:rewriting} for the templates used in paraphrasing. We have made this subset publicly available to facilitate community evaluation of compositionality.

The \textit{Private} subset comprises $155$ problem triplets, and the evaluation results presented in this paper are based on this subset. This portion of the data will not be made public to mitigate the risk of data leakage. \footnote{Researchers and institutions interested in evaluating their models on the \textit{Private} subset can contact us to arrange for testing.}
See Appendix \ref{dataset_info} for more annotation details about \textsc{MathTrap}.

\noindent\textbf{Problem Topic Composition}: \textsc{MathTrap} comprises problems from two main sources: $15.5\%$ from the GSM8K dataset and $84.5\%$ from the MATH dataset. It covers a diverse range of mathematical topics, including algebra ($23.2\%$), counting and probability ($22.6\%$), geometry ($16.1\%$), prealgebra ($12.3\%$), number theory ($7.74\%$), and precalculus ($2.58\%$).

\noindent\textbf{Trap Categories}: We carefully designed five categories of traps for constructing the \textsc{MathTrap} dataset. These include: 1) Concept Undefined: The reasoning process involves undefined mathematical concepts (such as $\tan90^\circ$, $0$ as a divisor, etc.). 2) Missing Condition: Lacking the necessary conditions required to solve the problem. 3) Direct Contradiction: Two conditions in the problem description directly contradict each other, which can be discovered without complex calculations. 4) Indirect Contradiction: There are indirect contradictions in the problem description, which can only be discovered during the reasoning process. 5) Violating Common Sense: The condition or final answer violates common sense.

\subsection{Evaluation Protocol}

%Open-source models consist of Llemma-MetaMath-7b trained on the Llemma mathematical foundation model \cite{azerbayev2024llemma}, as well as MetaMath-7b, MetaMath-13b, and MetaMath-70b \cite{yu2023metamath} obtained directly through SFT on Llama foundation models of varying scales.

We use \texttt{accuracy} as the evaluation metric for the problem. To measure compositional generalization, we calculate the \texttt{ratio} between the accuracy on trap questions and the accuracy on original questions. A model demonstrating good compositional generalization should exhibit similar performance before and after the introduction of trap questions, rather than showing degradation.

For Original Problems, following prior work \cite{yu2023metamath,gou2024tora,xi2024training,he2024self}, we calculate accuracy by determining whether the model’s final answer matches the standard answer. For Trap Problems and Conceptual Problems, we additionally check the intermediate steps of the model’s response to determine whether it correctly identified the trap. We employ GPT-4 as the judge for the checks. In experiments where GPT-4 might exhibit bias, we supplement our evaluation with results using Claude-3.5-Sonnet as an additional judge. We provide the prompts used for the evaluation in Table \ref{app_tab:judge} in the Appendix.

%We provide examples of the LLMs' correct and incorrect responses to trap problems and conceptual problems in Tables \ref{app_tab:output_of_ChatGPT4_trap_1}-\ref{app_tab:output_of_ChatGPT4_original_2} in the appendix.

\section{Results and Disscussion}
\subsection{The Compositionality of LLMs}
We evaluate the compositionality of LLMs on the \textsc{MathTrap} dataset, with results shown in Table \ref{tab:main}. %Note that for Conceptual Problems, we require the model to provide the correct reasoning behind its judgment, so the accuracy of random guessing is 0\% instead of 50\%. 
Proprietary LLMs achieve over $70\%$ accuracy on these Conceptual Problems, with OpenAI o1 even reaching $96.2\%$. This indicates that LLMs possess the knowledge required to identify most traps. However, when comparing LLM performance on original versus trap problems, we observe a significant decline. Most proprietary LLMs achieve less than half their original accuracy on trap problems. This indicates that even advanced, well-aligned LLMs struggle to apply trap-related knowledge flexibly to novel reasoning paths. Notably, o1-preview(Web) achieved a ratio of $77.4$, significantly higher than GPT-4's $51.2$. This suggests that \textit{o1's test-time scaling, akin to human `slow thinking', effectively improves LLM compositionality.} However, it still falls short of the human ratio of $85.9$. For detailed information on the human evaluation, please refer to Section \ref{sec:human_eval}.

%Furthermore, the open-source MetaMath series models achieve approximately 40\% accuracy on both concept and original problems, 
%but their performance significantly drops on trap problems compared to this figure.
%Looking at the ratio of accuracy between trap and original problems, open-source models all fall below 20\%. 

Open-source models' ratios below $20$ indicate that \textit{focusing solely on GSM8K and MATH accuracy doesn't truly enhance LLM reasoning abilities.}
Nevertheless, extensive pre-training (Llemma-MetaMath-7b vs. MetaMath-7b) and larger model scales (from 7b to 70b) still yield better compositional generalization effects (ratio increasing from $5.84\%$ to $19\%$).

%Furthermore, for the open-source MetaMath model series, while their accuracy on Conceptual problems is lower than closed-source commercial models, they can still correctly identify around 40\% of the traps. But when processing trap problems, their accuracy plummets to around 5\%. The accuracy ratios are all below 20\%.
%This implies that neither advanced closed-source commercial models nor open-source models can flexibly apply trap-related knowledge during the reasoning process.

\begin{table}[t]
\centering
\resizebox{\columnwidth}{!}{
\begin{tabular}{lcccc}
\toprule
\textbf{Model} & \textbf{Conceptual} & \textbf{Original} & \textbf{Trap} & \textbf{Ratio}\\ \hline
Gemini-Pro             & 70.0 & 36.9 & 8.30 & 22.5\\
Claude3-Opus           & 87.7 & 68.5 & 19.0 & 27.7\\
Claude-3.5-Sonnet       & 93.9 & 75.0 & 19.4 & 25.9\\
GPT-3.5-turbo-0125       & 74.6 & 40.5 & 7.60 & 18.8\\
GPT-4-0125-preview     & 90.0 & 70.3 & 36.0 & 51.2\\
o1-preview(API)        & 96.2 & 88.3 & 38.1 & 43.1 \\
o1-preview(Web)        & 92.3 & 87.5 & 67.7& 77.4 \\
Kimi                   & 71.5 & 46.1 & 19.6 & 42.5\\
\cmidrule{1-5}
Llemma-MetaMath-7B     & 55.2 & 41.4 & 6.40 & 15.5\\
MetaMath-7B            & 43.2 & 32.5 & 1.90 & 5.84\\
MetaMath-13B           & 37.8 & 37.5 & 3.90 & 10.4\\
MetaMath-70B           & 57.6 & 34.2 & 6.50 & 19.0\\ 
Llama3-8B              & 70.5 & 33.3 & 6.45 & 19.4\\ 
Llama3-8B-Base         & 44.7 & 33.3 & 6.45 & 19.4\\ 
Llama3-70B             & 88.5 & 61.7 & 7.74 & 12.5\\ 
Llama3-70B-Base        & 53.8 & 37.5 & 7.74 & 20.6\\
Llama3.1-8B            & 70.8 & 61.7 & 13.5 & 21.9\\
Llama3.1-70B           & 88.5 & 69.2 & 19.4 & 28.0\\
\bottomrule
\end{tabular}
}
\caption{Accuracy ($\%$) of various models on three types of \textsc{MathTrap} problems. `Conceptual' represents Conceptual problems, `Original' refers to the original problems, and `Trap' denotes the trap problems. `Ratio' refers to the ratio of the accuracy on Trap problems to the accuracy on Original problems. It reflects the degree to which the performance is maintained when facing problems with traps, relative to the original problems.} %For `T of F' problems, the models need to provide the correct reasoning behind its judgement. Therefore, the accuracy for random guessing is not 50.0 but rather 0.
\label{tab:main}
\end{table}
\begin{table}[t]
\centering
\resizebox{0.7\linewidth}{!}{
\begin{tabular}{@{}lc@{}}
\toprule
\textbf{Condition} & \textbf{Human Accuracy } \\ \midrule
Trap Problem(w/o Notice) & 83.8\\
Trap Problem(w/ Notice) & 95.1\\
Original Problem & 97.6\\ \bottomrule
\end{tabular}
}
\caption{Human accuracy(\%) on \textsc{MathTrap}. ``Trap Problem (w/o Notice)'' refers to the accuracy of human solutions when unaware that the problems contain traps. ``Trap Problem (w/ Notice)'' indicates the accuracy of human solutions when informed that the problems contain traps. ``Original Problem'' refers to the accuracy of human solutions on the original problems.}
%Human accuracy across different problem settings. ``Trap Problem'' refers to the accuracy of human responses when encountering the Trap Problem for the first time without any prompts. ``Trap Problem with Prompt'' denotes the accuracy after humans are alerted to the potential presence of a trap. ``Original Problem'' represents the accuracy of responses to standard problems.}
\label{tab:human}
\end{table}

\begin{table*}[t]
\centering
\resizebox{\textwidth}{!}{
\begin{tabular}{l|l|cccc|cccc}
\toprule
\multicolumn{1}{l|}{\textbf{Model}}&\multicolumn{1}{l|}{\textbf{Judge}} & \multicolumn{4}{c|}{\textbf{Original Problem}} & \multicolumn{4}{c}{\textbf{Trap Problem}} \\
 && \textbf{w/o Notice} & \textbf{w/ Notice} & \textbf{ICL (1-shot)} & \textbf{ICL (5-shot)} & \textbf{w/o Notice} & \textbf{w/ Notice} & \textbf{ICL (1-shot)} & \textbf{ICL (5-shot)} \\
\midrule
Gemini-pro& GPT-4 & 36.9 & 37.8 & 32.4 & 50.0 & 8.3 & 14.1 & 8.94 & 27.7 \\
Claude3-Opus& GPT-4 & 68.5 & 65.8 & 68.5 & 82.5 & 19.0 & 40.7 & 29.0 & 56.1 \\
Claude3-Opus& Claude-3.5 & 67.6 & 64.0 & 68.5 & 70.8 & 16.1 & 48.4 & 23.9 & 47.1 \\
GPT-3.5-turbo-0125& GPT-4 & 40.5 & 40.5 & 45.9 & 51.7 & 7.74 & 12.2 & 12.2 & 23.9 \\
GPT-3.5-turbo-0125 &Claude-3.5 & 37.8 & 29.7 & 44.1 & 46.7 & 7.10 & 13.5 & 11.0 & 12.9 \\
GPT-4-0125-preview& GPT-4 & 70.3 & 72.1 & 65.8 & 77.5 & 35.5 & 50.3 & 41.9 & 48.4 \\
GPT-4-0125-preview&Claude-3.5 & 65.8 & 67.6 & 64.9 & 76.7 & 24.5 & 40.6 & 36.8 & 35.5 \\
Kimi& GPT-4 & 46.1 & 41.4 & 41.4 & 55.0 & 19.6 & 37.1 & 26.4 & 31.0 \\
\cmidrule{1-10}
Llemma-MetaMath-7B& GPT-4 & 41.4 & 42.7 & 32.2 & 33.3 & 6.36 & 9.86 & 7.93 & 7.74 \\
MetaMath-7B& GPT-4 & 32.5 & 32.5 & 30.0 & 35.0 & 1.94 & 3.23 & 5.16 & 17.4 \\
MetaMath-13B& GPT-4 & 37.5 & 37.5 & 34.2 & 40.7 & 3.87 & 3.87 & 2.58 & 12.3 \\
MetaMath-70B& GPT-4 & 34.2 & 30.8 & 35.8 & 38.4 & 6.45 & 7.74 & 4.52 & 16.8 \\
Llama3-8B& GPT-4 & 33.3 & 36.7 & 38.3 & 35.8 & 6.45 & 13.5 & 4.52 & 35.5 \\
Llama3-8B-Base& GPT-4 & 14.2 & 16.7 & 15.8 & 24.2 & 3.23 & 3.87 & 5.16 & 20.0 \\
Llama3-70B& GPT-4 &61.7 & 61.7 & 54.2 & 61.7 & 7.74 & 26.5 & 14.8 & 39.4 \\
Llama3-70B-Base& GPT-4 & 37.5 & 30.0 & 34.2 & 41.7 & 7.74 & 9.03 & 7.74 & 25.8 \\
\bottomrule
\end{tabular}
}
%prompt代表直接提示模型可能存在unreasonable query
\caption{The impact of external intervention methods on the accuracy for original problems and trap problems. ``w/o Notice'' refers to the control experiment without any external intervention. 'w/ Notice' indicates using a natural language prompt to inform the model that the problem description may contain traps. ICL (1/5-shot) refers to adding one or five demonstrations in the context to exemplify how to handle trap problems. The prompt templates employed are presented in Tables \ref{app_tab:no_solution}-\ref{app_tab:5_shot} in the Appendix.}
\label{tab:intervention}

\end{table*}

%在不同数据集的混合方式下对Llemma模型进行微调，在Original Problem和Trap Problem上的准确率
\begin{table}[t]
\centering
\resizebox{\columnwidth}{!}{
\begin{tabular}{lccc}
\toprule
\textbf{Dataset} & \textbf{Original} & \textbf{Trap(GPT-4)}& \textbf{Trap(Claude-3.5)} \\ \midrule
GSM8K+MATH & 20.8 & 1.01  &0.65\\
GSM8K+MATH+MathTrap1K & 13.3 & 12.4 &34.2 \\
MetaMath395K & 41.4 & 6.36 &1.94\\
MetaMath395K+MathTrap1K & 33.3 & 29.1 &11.6        \\ \bottomrule
\end{tabular}
}
\caption{The impact of fine-tuning data configurations on the accuracy for original and trap problems. We use Llemma as the foundation model. The parentheses indicate the judge model used.}

\label{tab:fine_tune}
\end{table}

\subsection{The Compositionality of Human}
\label{sec:human_eval}
As a control experiment, we evaluate human performance on the \textsc{MathTrap} dataset. Specifically, we recruit 43 undergraduate students majoring in science and engineering from top universities to participate in the experiment. Each student is randomly assigned two problems: one original problem and one trap problem.\footnote{We only assign two problems because once participants discover that a problem contain a trap, they would consciously look for traps in subsequent problems, which would affect the accuracy of the results.} During the answering process, participants are not aware that the assigned problems might contain traps. To prevent participants from discovering the traps by comparing the two problems, the original problem and the trap problem assigned to each participant are completely different. The results are shown in Table \ref{tab:human}. Humans achieve an accuracy of 83.8\% on the trap problems. \textit{In terms of the ratio of accuracy on trap problems to original problems, humans attained an accuracy ratio of $83.8/97.6=85.9\%$, far surpassing all existing LLMs}. This indicates that humans demonstrate strong compositional reasoning ability on the \textsc{MathTrap} dataset. For cases where participants fail to correctly identify the traps, we further inform them that the problems might contain traps and ask them to answer again. \textit{After receiving this hint, the human accuracy rate increases from 83.8\% to 95.1\%, almost identical to their performance on the original problems.}

\subsection{Mitigating LLMs' Failure on MathTrap}

The results in Table \ref{tab:main} show that LLMs have acquired the relevant knowledge needed to solve trap problems, but are unable to spontaneously apply this knowledge to reasoning on trap problems. Therefore, we attempt to mitigate this issue through external interventions, with the specific results presented in Table \ref{tab:intervention} and \ref{tab:fine_tune}. The GPT series models constitute a significant subset of proprietary models in Table \ref{tab:intervention}. To mitigate potential bias in evaluating these models using GPT-4, we supplement our assessment with results using Claude-3.5-Sonnet as a judge.

\noindent\textbf{Natural language prompt}: Adding the prompt ``Note that this problem may be unsolvable or has no solution'' before the problem statement. Refer to Table \ref{app_tab:no_solution} for the full prompt. Our results show that natural language prompts can guide LLMs to notice the contradictions or traps in the problem descriptions without affecting the accuracy on ``Original Problems'', especially for those well-aligned closed-source LLMs.

\noindent\textbf{Few-shot demonstration}: In the 1-shot setting, a randomly sampled trap problem and its reasoning process are inserted into the context (refer to Table \ref{app_tab:1_shot} for the full prompt); in the 5-shot setting, 2 original problems and 3 trap problems with their reasoning processes are inserted (refer to Table \ref{app_tab:5_shot}). The results show that compared to natural language prompts, few-shot demonstrations are more effective in handling trap problems. Additionally, in the 5-shot setting with a mix of original problems, the accuracy on original problems is also improved.

\noindent\textbf{Fine-tuning}: We use the \textsc{MathTrap} public subset containing $1,000$ problem triplets for fine-tuning. Given that GPT-4 was employed for data augmentation, we included additional evaluation results using Claude-3.5-Sonnet in Table 5 to avoid potential assessment bias from using GPT-4 as both an augmenter and judge. The experiments demonstrate that fine-tuning can significantly improve model performance on trap problems without prompt, but it may also reduce the accuracy of solving original problems.

\section{Conclusions}
This paper investigates the compositional generalization of LLMs in mathematical reasoning. By introducing traps into the problem descriptions, we construct novel ``Trap Problems'' that cannot be solved by merely following trained reasoning paths for LLMs. Experiments on \textsc{MathTrap} demonstrate that LLMs fail to spontaneously combine their learned knowledge to reason about trap problems. Although this limitation can be mitigated through external interventions, there remains a significant gap compared to the compositional generalization capabilities of humans.

\section*{Limitations}
The construction of trap problems places high demands on the annotators' abilities, resulting in high annotation costs. Therefore, how to automatically generate high-quality trap problems through automated methods is worthy of further investigation.

\bibliography{custom}
\appendix
\section*{Appendices} 
\section{Related Works}
\subsection{Investigation on the Limitations of Transformer Capabilities}
\label{app:limitations}
In recent years, large language models (LLMs) have achieved unprecedented success in various tasks requiring complex reasoning \cite{openai2023gpt4}, such as coding \cite{guo2024deepseekcoder,zheng2024opencodeinterpreter} and solving mathematical problems \cite{luo2023wizardmath,toshniwal2024openmathinstruct1}. Some researchers even view these exciting advancements as sparks of artificial general intelligence \cite{bubeck2023sparks}. In stark contrast, these models have shown unexpected failures in simple and intuitive tasks \cite{bian2024chatgpt,koralus2023humans}. For example, the state-of-the-art GPT-4 only achieve 59\% accuracy on three-digit multiplication problems \cite{dziri2023faith}. 

What is the reason behind this stark discrepancy? Recent studies have examined LLMs' ability to composite knowledge from training scenarios to solve more complex problems. Tests covered tasks such as boolean variable assignment \cite{anil2022exploringlengthgeneralizationlarge}, semantic parsing \cite{hosseini-etal-2022-compositional}, deductive reasoning \cite{sanyal-etal-2022-robustlr}, and arithmetic reasoning \cite{kazemi-etal-2023-lambada}. A common trend shows that as problem complexity increases, LLMs' accuracy drops significantly. Our \textsc{MathTrap} is based on math word problems, currently the most widely studied task for evaluating LLM reasoning. Unlike previous work that assessed models' compositional generalization by increasing problem complexity, we simply introduced straightforward logical traps into original problems without significantly increasing their complexity.

Researchers have also investigated whether LLMs' impressive reasoning abilities stem from learning general knowledge or merely reciting examples from their vast training corpora. Studies \cite{wu-etal-2024-reasoning,zhang2023counterfactual} on LLMs' performance in factual versus counterfactual scenarios revealed significant performance drops in counterfactual cases, suggesting LLMs often recite answers from common cases in their training data. Recent research \cite{dziri2023faith} modeled reasoning tasks as computational graphs, experimentally demonstrating that LLMs reason through subgraph matching rather than developing systematic problem-solving skills. Another study \cite{hu2024casebasedrulebasedtransformersmath} removed certain samples from the training set and found that LLMs rely on surrounding cases in the training set for mathematical reasoning rather than learning generalizable rules. These findings indicate that LLMs still face challenges in learning knowledge and combining them to solve out-of-distribution problems. 

\subsection{Math Word Problem Benchmark}
Mathematical word problems have long been considered an effective proxy for evaluating the reasoning abilities of large language models (LLMs), garnering widespread attention from the academic community. Numerous benchmark datasets of math word problems have been proposed. ASDiv \cite{miao-etal-2020-diverse} is a dataset covering most knowledge types encountered in elementary school, with each problem annotated with its corresponding knowledge component and grade level. SVAMP \cite{patel-etal-2021-nlp} comprises 1,000 challenging elementary math word problems carefully designed and curated to assess a system's complex reasoning and multi-step arithmetic capabilities. \citet{DBLP:journals/corr/abs-2110-14168}  introduced GSM8K, a high-quality and diverse evaluation benchmark containing 8.5k math word problems. \citet{DBLP:journals/corr/abs-2103-03874} presented MATH, a dataset of 12,500 challenging competition math problems. Recently, researchers have also constructed unanswerable math word problems to evaluate hallucinations in LLMs' mathematical reasoning process. However, evaluating LLMs' compositional abilities has been limited to symbolic reasoning tasks \cite{article,dziri2023faith}. \citet{tang2024on} found that semantics play a crucial role in the reasoning process, motivating our \textsc{MathTrap} dataset to evaluate LLMs' compositional skills on semantically rich math word problems.
%错题评测	Following is a problem with no solution or can't be solved, and a reference answer about how to find the contradiction. There is also an answer you need to make a verdict on whether it finds the same contradiction as the reference answer does. Please output a short comment and end with [0] or [1] ([1] means the answer finds the same contradiction and explicitly states it). #The Problem: {input} \n#The Reference Answer: {ref}\n#The Answer Needs Verdict: {answer}\n#Your Verdict: 

\section{Annotation Process and Standards of \textsc{MathTrap} Dataset}
\label{dataset_info}
In this section, we provide a detailed explanation covering three aspects: the annotators' background, annotation criteria, and the annotation process.
\subsection{Qualified annotators}
Our annotation team consists of five students with STEM backgrounds from top universities, with an average math score of $145$ out of $150$ in their college entrance exams. They possess the necessary math knowledge and passed our qualification test, ensuring their ability to understand and complete our tasks with high quality.

\subsection{Clear and Specific Annotation Criteria}
\begin{itemize}
    \item \textit{Adherence to trap problem definition}: We provided clear definitions and examples for five types of traps (as shown in \ref{app_sec:category}). Annotators were required to fully understand these definitions and verify that their trap problems align with the corresponding trap definitions.
    \item \textit{Unambiguity}: We require annotators to mutually verify the semantic clarity of trap problems, ensuring the modified questions are unambiguous.
    \item \textit{Difficulty}: We set a standard that the trap problems should cause GPT-3.5 to give incorrect responses in at least 1 out of 5 runs, preventing overly simple trap problems.
    \item \textit{Diversity}: We dynamically monitored the distribution of knowledge points during the annotation process and provided feedback to annotators to adjust their selection of topics.
\end{itemize}

\subsection{Standardized Annotation Process}
\begin{itemize}
    \item \textit{Problem assignment and distribution}: Problems from each knowledge point in the original datasets were equally distributed among five annotators. This approach was adopted to prevent potential bias that might arise if any single annotator were to focus on only a subset of knowledge points.
    \item \textit{Quality control during annotation}: We required strict adherence to annotation standards and recording of the verification process. A supervisor regularly checked the output for compliance with our standards and provided feedback.
    \item \textit{Post-annotation quality assessment}: All annotators cross-verified trap problems created by others, filtering out those that didn't meet the annotation standards.
\end{itemize}
\section{Evaluation}
\label{app:protocol}
\subsection{Compared Method} We evaluate mainstream proprietary and open-source LLMs with strong mathematical capabilities. Proprietary models include Claude3-Opus, Gemini-pro, OpenAI o1, GPT-4-0125-preview, GPT-3.5-turbo-0125, and Kimi. The open-source models include the 7B, 13B, and 70B versions of MetaMath \cite{yu2023metamath}, which are all based on the Llama-2 model. Additionally, we provide Llemma-MetaMath-7B, which is based on Llemma \cite{azerbayev2024llemma}, a foundation model pretrained specifically for mathematics.

\subsection{Prompt Template}
%在表6展示了使用GPT-4测评各个模型回答准确率的提示词模板，在表7展示了使用GPT-4进行训练集答案增强的提示词模板，在表8展示了提示(Notice)模型问题可能无解使用的提示词模板，在表9展示了1-shot设置下in context learning的提示词模板，在表10展示了5-shot设置下in context learning的提示词模板。
Table \ref{app_tab:judge} presents the prompt templates used to evaluate the accuracy of responses from various models using GPT-4. Table \ref{app_tab:rewriting} displays the prompt templates for enhancing the training set answers with GPT-4. Table \ref{app_tab:no_solution} shows the prompt templates used to indicate that a model's question might be unsolvable. Table \ref{app_tab:1_shot} presents the prompt templates for in-context learning in a 1-shot setting. Table \ref{app_tab:5_shot} presents the prompt templates for in-context learning in a 5-shot setting.

% \subsection{GPT-4 as a judge}
% \label{app:gpt_as_a_judge}
% We employed GPT-4 as the judge for the above evaluation process. We manually inspected $50$ of GPT-4's judgments and found only $3$ cases where its assessment was inaccurate. Therefore, GPT-4 has the ability to judge whether an answer is correct or not. 

\begin{table*}[ht]
\centering
\begin{tabular}{@{} p{\textwidth} @{}}
\toprule
Following is a problem with no solution or can't be solved, and a reference answer about how to find the contradiction. There is also an answer you need to make a verdict on whether it finds the same contradiction as the reference answer does. Please output a short comment and end with [0] or [1] ([1] means the answer finds the same contradiction and explicitly states it). \\
\textbf{\#The Problem: } \{input\} \\
\textbf{\#The Reference Answer: } \{ref\} \\
\textbf{\#The Answer Needs Verdict: } \{answer\} \\
\textbf{\#Your Verdict: }  \\
\bottomrule
\end{tabular}
%使用GPT-4对各个LLMs进行Trap Problem的评测时使用的Prompt template
\caption{Prompt template used for evaluating the Trap Problem across various Large Language Models (LLMs) using GPT-4.}
\label{app_tab:judge}
\end{table*}

%改写问题	Following is a problem with no solution or can't be solved, and a brief comment on why it has no solution. Please generate an answer that try to solve the question, until you meets where comment says it's unable to solve, and explain the situation more informatively. The answer should not mention there exists "comment", you should pretend to find the contradiction independently. Question: "{question}" Comment: "{comment}"

\begin{table*}[ht]
\centering
\begin{tabular}{@{} p{\textwidth} @{}}
\toprule
Following is a problem with no solution or can't be solved, and a brief comment on why it has no solution. Please generate an answer that try to solve the question, until you meets where comment says it's unable to solve, and explain the situation more informatively. The answer should not mention there exists ``comment'', you should pretend to find the contradiction independently.   \\
\textbf{\#Question: } \{question\} \\
\textbf{\#Comment: } \{comment\} \\
\textbf{\#Rephrased Answer: }  \\
\bottomrule
\end{tabular}
%使用GPT-4对Trap Problem进行问题改写使用的Prompt template，同时保持改写的问题和原来问题涉及的Trap类型一致
\caption{Prompt template used for answer augmentation of the Trap Problem using GPT-4.}
\label{app_tab:rewriting}
\end{table*}

%prompt of trap Please solve the following problem. Note that this problem may be unsolvable or has no solution.\n\n### Instruction:\n{instruction}\n\n### Response: Let's think step by step.
\begin{table*}[ht]
\centering
\begin{tabular}{@{} p{\textwidth} @{}}
\toprule
Please solve the following problem. Note that this problem may be unsolvable or has no solution. \\
\textbf{\#\#\# Instruction: }\\ \{instruction\} \\
\textbf{\#\#\# Response: }  \\
Let's think step by step. \\
\bottomrule
\end{tabular}
%直接提示Trap Problem可能无解的Prompt template
\caption{The prompt template uesd for directly suggesting to large language models (LLMs) that the problem might be unreasonable.}
\label{app_tab:no_solution}
\end{table*}
%prompt for 1-shot Please solve the following problem. Note that this problem may be unsolvable or has no solution.\n"
%"Here is an example: ### Instruction:\nIf $r$ is a negative integer, and the maximum value of the expression $5r^2 + br - 12$ is 36, what's the value of b? \n\n### Response: Let's think step by step. A parabola opening upwards doesn't have a maximum value for negative integer n. The answer is \\boxed{{\\text{{undefined}}}}."
%"### Instruction:\n{instruction}\n\n### Response: Let's think step by step."
\begin{table*}[ht]
\centering
\begin{tabular}{@{} p{\textwidth} @{}}
\toprule
Please solve the following problem. Note that this problem may be unsolvable or has no solution. \\
Here is an example: \\
\textbf{\#\#\# Instruction: }\\
If $r$ is a negative integer, and the maximum value of the expression $5r^2 + br - 12$ is 36, what's the value of b? \\
\textbf{\#\#\# Response: }Let's think step by step. A parabola opening upwards doesn't have a maximum value for negative integer r. \\
\textbf{\#\#\# Instruction: }  \\
\{instruction\} \\
\textbf{\#\#\# Response: }Let's think step by step.\\
\bottomrule
\end{tabular}
%使用的包含1个例子作为示例的Prompt template
\caption{The prompt template used under the 1-shot setting for in-context learning.}
\label{app_tab:1_shot}
\end{table*}

\begin{table*}[ht]
\centering
\begin{tabular}{@{} p{\textwidth} @{}}
\toprule
Please solve the following problem. Note that this problem may be unsolvable or has no solution. \\
Here is an example: \\
\textbf{\#\#\# Instruction: }\\
If $r$ is a negative integer, and the maximum value of the expression $5r^2 + br - 12$ is 36, what's the value of b? \\
\textbf{\#\#\# Response: }Let's think step by step.\\
...\\
Regardless of whether $r$ is positive or negative, a parabola that opens upwards does not have a maximum value.\\Therefore, it is impossible to equate the maximum value of the expression to 36 as the problem initially asks, and we cannot solve for $b$ under these conditions. The question has posed an unsolvable scenario.\\The problem is unsolvable. \\
\textbf{\#\#\# Instruction: }\\
How many prime numbers can be divided by 39? \\
\textbf{\#\#\# Response: }Let's think step by step. No prime numbers can be divided by 39.\\
...\\
Therefore, there are \boxed{0} prime numbers that can be divided by 39. \\The answer is: 0\\ \\
\textbf{\#\#\# Instruction: }\\
If odd integers $x y$ satisfies $\frac{1}{2}-\frac{1}{3}=\frac{x}{y}$, then what is the value of $x$? \\
\textbf{\#\#\# Response: }Let's think step by step. To solve this problem, we must first simplify the left-hand side of the equation, $\frac{1}{2}-\frac{1}{3}$. This simplifies to $\frac{1}{6}$.  \\
...\\
Therefore, under the given conditions, there is no solution to this problem. The constraints on $x$ and $y$ being odd integers and the equation $\frac{1}{2}-\frac{1}{3}=\frac{x}{y}$ cannot hold true simultaneously.\\The problem is unsolvable. \\

\textbf{\#\#\# Instruction: }\\
If a triangle has two sides of lengths 5 and 7 units, then how many different integer lengths can the third side be? \\
\textbf{\#\#\# Response: }Let's think step by step. Let the two given sides be $a$ and $b$.\\
... \\
In this case, $a=5$ and $b=7$, so the inequalities become $|5-7| < c < 5+7$, which simplify to $2 < c < 12$.\\Since $c$ must be an integer, the possible integer lengths for the third side are 3, 4, 5, 6, 7, 8, 9, 10, and 11.\\Therefore, there are $\boxed{9}$ different integer lengths the third side can be.\\The answer is: 9\\
\textbf{\#\#\# Instruction: }\\
Find p if the focus of the parabola $x + x^2 = -\frac{p}{12} y^2.$ is (-1/2,1).\\
\textbf{\#\#\# Response: }Let's think step by step. To begin solving this problem, we first need to express the equation in the standard form of a parabola.  \\
... \\
Therefore, we can't express the equation in the standard form of a parabola, which indicates that the equation does not represent a parabola. Consequently, we cannot find a focus for a parabola because the given equation does not describe a parabola. The initial assumption that we are dealing with a parabola is incorrect.\\The problem is unsolvable. \\
\textbf{\#\#\# Instruction: }  \\
\{instruction\} \\
\textbf{\#\#\# Response: }Let's think step by step.\\
\bottomrule
\end{tabular}
%In context learning 5-shot的设置下的Prompt template
\caption{The prompt template used under the 5-shot setting for in-context learning.}
\label{app_tab:5_shot}
\end{table*}

\begin{table*}[h]
    {  \footnotesize \begin{tabularx}{\linewidth}{cXX} 
    \toprule 
        \makecell[c]{\textbf{Type}} &\makecell[c]{\textbf{Trap problem}} & \makecell[c]{\textbf{Explanation}}\\ \midrule
    \textbf{Concept Undefined}&{In right triangle $XYZ$ with $\angle YXZ = 90^\circ$, we have $XY = 24$ and $YZ = 25$. Find  \colorbox {yellow}{$\tan X$}.}
    &$\angle YXZ = 90^\circ$, so $\tan X=\tan(\pi/2)$ is undefined.\\
    \hline
   \textbf{Missing Condition}& Natalia sold 48 clips in April and half as many clips in May. How many clips did Natalia sell altogether in April and  \colorbox {yellow}{June}?& {We don't know anything about June, so it's impossible to calculate the sum of the sales for April and June.}\\
    \hline
    \textbf{Direct Contradiction}& {An equilateral triangle has a perimeter of 30 centimeters\colorbox {yellow}{and a height of 10 centimeters}. Calculate the area of the triangle.}& {The height of the equilateral triangle and its side length are both 10 centimeters, which is contradictory and impossible.}\\
    \hline
    \textbf{Indirect Contradiction}& {Find the \colorbox {yellow}{integer} solution of the equation $x^2+x=3$.}& {The 2 solutions of this quadratic equation is $\frac{-1 \pm \sqrt{13}}{2}$}, so there is no integer solution.\\
    \hline
    \textbf{Violating Common Sense}& {Max picks \colorbox {yellow}{5} different cards without replacement from a standard 52-card deck. What is the probability that the cards are of different suits?}& {There are only 4 suits in a deck, so it's impossible for 5 cards to be of different suits.}\\
  \bottomrule 
 \end{tabularx} }
 %对每个类别的Trap Problem例子的解释
    \caption{Explanation of examples of trap problems for each category. The sections highlighted in \colorbox {yellow}{yellow} delineate the distinction between original problems and trap problems.} 
    \label{app_tab:explation}
\end{table*}

\begin{table*}
    % \centering
    \setlength{\tabcolsep}{2pt}
    {
    \renewcommand\arraystretch{1.0}
    \resizebox{\linewidth}{!}{
        \begin{tabular}{|m{4cm}|m{6cm}|m{7.5cm}|}
        % \toprule
        \hline
        \makecell[c]{\textbf{Trap problem}} & \makecell[c]{\textbf{Output(w/o Notice)}} & \makecell[c]{\textbf{Output(w/ Notice)}}  \\
        \hline
In right triangle $XYZ$ with $\angle YXZ = 90^\circ$, we have $XY = 24$ and $YZ = 25$. Find  \colorbox {yellow}{$\tan X$}.
    &
Alright, in our right triangle $XYZ$ where $\angle YXZ = 90^\circ$, we're given the lengths $XY = 24$ and $YZ = 25$. To find $\tan X$, we're essentially looking for the ratio of the opposite side to the adjacent side of angle $X$. Let's break it down:
\par...\par
% 1. **Identify the sides**: Since $XY$ is one of the sides and $YZ$ is the hypotenuse (the longest side in a right triangle), we're missing the length of $XZ$, which is the side adjacent to angle $X$.2. **Use the Pythagorean theorem**: This theorem tells us that in a right triangle, the square of the hypotenuse is equal to the sum of the squares of the other two sides. Mathematically, it's expressed as $c^2 = a^2 + b^2$, where $c$ is the length of the hypotenuse, and $a$ and $b$ are the lengths of the other two sides. Here, we can plug in our values with $YZ$ as the hypotenuse ($c = 25$) and $XY$ as one of the sides ($a = 24$), and solve for $b$ (the length of $XZ$).3. **Calculate $XZ$**: Plugging in our values, we get $25^2 = 24^2 + XZ^2$. Solving this will give us the length of $XZ$.4. **Find $\tan X$**: Remember, $\tan X = \frac{\text{opposite}}{\text{adjacent}}$ in the context of angle $X$. So, once we have the length of $XZ$, we can calculate $\tan X = \frac{XY}{XZ}$.
Let's go through the calculation:- First, solve for $XZ^2$: $25^2 = 24^2 + XZ^2$ leads to $625 = 576 + XZ^2$, which simplifies to $XZ^2 = 625 - 576 = 49$.- Taking the square root of both sides gives $XZ = 7$.Finally, to find $\tan X$, we use the lengths $XY$ (opposite side to $X$) and $XZ$ (adjacent side to $X$): $\tan X = \frac{24}{7}$.So, \colorbox {red}{$\tan X = \frac{24}{7}$}. This gives us a detailed step-by-step approach to finding $\tan X$ in the triangle $XYZ$ where $\angle YXZ = 90^\circ$, with $XY = 24$ and $YZ = 25$.
    &
    Given a right triangle $XYZ$ with $\angle YXZ = 90^\circ$, we're tasked with finding $\tan X$. We have the lengths of two sides: $XY = 24$ and $YZ = 25$. The side $XY$ can be considered as the side opposite to $\angle YXZ$, and $YZ$ can be thought of as the hypotenuse of the triangle.However, to find $\tan X$, we actually need to know the relationship between the side adjacent to $\angle X$ (which would be $XZ$ in this case) and the side opposite to $\angle X$ (which is $XY$). 
    \par...\par
    % Remember, $\tan$ of an angle in a right triangle is the ratio of the length of the side opposite the angle to the length of the side adjacent to the angle. The formula is $\tan(\theta) = \frac{\text{opposite}}{\text{adjacent}}$.Given that $XY = 24$ is the side opposite to $\angle X$ and $YZ = 25$ is the hypotenuse, we can use the Pythagorean theorem to find the length of $XZ$, the side adjacent to $\angle X$. The Pythagorean theorem states that in a right triangle, the square of the length of the hypotenuse ($c$) is equal to the sum of the squares of the lengths of the other two sides ($a$ and $b$): $a^2 + b^2 = c^2$.
    Let's apply this theorem to find $XZ$:$$XY^2 + XZ^2 = YZ^2$$Plugging in the values we know, we get:$$24^2 + XZ^2 = 25^2$$Solving for $XZ^2$, we get:$$XZ^2 = 25^2 - 24^2 = 625 - 576 = 49$$Taking the square root of both sides gives us $XZ = 7$.Now that we have the length of the side adjacent to $\angle X$, we can find $\tan X$:$$\tan X = \frac{XY}{XZ} = \frac{24}{7}$$Therefore, \colorbox {red}{$\tan X = \frac{24}{7}$}.
    \\
    \hline
    Natalia sold 48 clips in April and half as many clips in May. How many clips did Natalia sell altogether in April and  \colorbox {yellow}{June}?&
    Let's break it down:1. **April Sales:** Natalia sold clips to 48 friends in April.2. **May Sales:** It's mentioned that she sold to only half the number of friends in April, which would be $\frac{1}{2} \times 48 = 24$ friends in May.However, to find the **cumulative number** of clips sold in April and **June**, we need a bit more information about June, as the instruction doesn't provide any details for June's sales.Based on the provided details, we can calculate the cumulative number for April and May, but without the June data, \colorbox {green}{we can't complete the request }as specified. If you have any information about June's sales or if there was a typo, and you meant to ask for the cumulative total for April and May, please let me know!
    &
        Alright, let's break this down step by step to understand the situation:1. **April Sales:** We know that Natalia managed to sell clips to 48 of her buddies in April.2. **May Sales:** It's stated that she sold half of that amount in May. So, if she sold 48 clips in April, half of that amount is $48 \div 2 = 24$ clips in May.3. **June Sales:** However, the instruction asks for the total number of clips sold in April and **June**, but we don't have any information about June's sales. The mention of May's sales seems to be there to potentially confuse or add extra information that doesn't directly pertain to the question as asked.Given that we only have data for April and May, and the question asks for the total for April and June without providing information on June's sales, we can't accurately determine the total number of clips Natalia sold in both April and June. Without the June sales figure, \colorbox {green}{the problem doesn't have a complete solution} based on the information provided.
\\
        \hline
        % \bottomrule
        \end{tabular}
        }
    
    \caption{Responses of GPT-4-0125-preview to Trap Problems. ``Output(w/o Notice)'' refers to the model's output when no additional prompt is provided, whereas ``Output(w/ Notice)'' denotes the outputs when the model is informed that the problem may be unreasonable. The sections highlighted in \colorbox {yellow}{yellow} delineate the distinction between original problems and trap problems. The \colorbox {green}{green} sections represent instances where the model's final answers are correct, while the \colorbox {red}{red} sections indicate where the model's final answers are incorrect.}
    \label{app_tab:output_of_ChatGPT4_trap_1}
    }
\end{table*}

\begin{table*}
    % \centering
    \setlength{\tabcolsep}{2pt}
    {
    \renewcommand\arraystretch{1.0}
    \resizebox{\linewidth}{!}{
        \begin{tabular}{|m{4cm}|m{6cm}|m{7.5cm}|}
        % \toprule
        \hline
        \makecell[c]{\textbf{Trap problem}} & \makecell[c]{\textbf{Output(w/o Notice)}} & \makecell[c]{\textbf{Output(w/ Notice)}}  \\
%         \hline Natalia sold 48 clips in April and half as many clips in May. How many clips did Natalia sell altogether in April and  \colorbox {yellow}{June}?&Sure, let's break it down.
% First, we know Natalia sold 48 clips in April.
% Then, in May, she sold half as many clips, which means 48/2=24 clips.
% To find out how many clips Natalia sold altogether in April and May, we add the number of clips she sold in April and May:
% 48+24=72.
% So, Natalia sold 72 clips altogether in April and May.& Certainly! First, let's find out how many clips Natalia sold in May. Since she sold half as many clips as in April, we'll divide the number of clips sold in April by 2:
% 48 (clips sold in April) ÷ 2 = 24 clips sold in May
% Now, let's calculate the total number of clips sold in April and May:
% April: 48 clips
% May: 24 clips
% Total clips sold in April and May: 48 (April) + 24 (May) = 72 clips
% So, Natalia sold a total of 72 clips in April and May. However, the question asks about clips sold in April and June. Since there's no mention of June, we can't determine how many clips she sold in June. Therefore, we can't find the total clips sold in April and June.\\
        \hline
        An equilateral triangle has a perimeter of 30 centimeters and \colorbox {yellow}{a height of 10 centimeters}. Calculate the area of the triangle.&
%         To find the area of an equilateral triangle, we use the formula:

% \[
% \text{Area} = \frac{1}{2} \times \text{base} \times \text{height}
% \]

% First, we calculate the side length of the equilateral triangle. Since the perimeter is 30 centimeters, and an equilateral triangle has three equal sides:

% \[
% \text{Side length} = \frac{30 \text{ cm}}{3} = 10 \text{ cm}
% \]

% Now, apply the area formula using the side length as the base and the given height of 10 centimeters:

% \[
% \text{Area} = \frac{1}{2} \times 10 \text{ cm} \times 10 \text{ cm} = 50 \text{ cm}^2
% \]

% Therefore, the area of the triangle is \(50 \text{ cm}^2\).
Given that the equilateral triangle has a perimeter of 30 centimeters, we can first find the length of one side. Since an equilateral triangle has all sides of equal length, we can divide the perimeter by 3:$$\text{Side length} = \frac{\text{Perimeter}}{3} = $$$$\frac{30}{3} = 10 \text{ cm}$$Now, to find the area of the triangle, we can use the formula for the area of a triangle, which is $\frac{1}{2} \times \text{base} \times \text{height}$. However, for an equilateral triangle, we can also use the formula involving its side length and height. Since we know both the side length and the height, we can use the following formula where \( h \) is the height:$$\text{Area} = \frac{1}{2} \times \text{Side length} \times h$$$$ = \frac{1}{2} \times 10 \times 10 = 50 \text{ cm}^2$$Therefore, the area of the equilateral triangle is \colorbox {red}{\( 50 \text{ cm}^2 \)}.
& 
% To solve this problem, let's start by confirming the parameters of an equilateral triangle given its properties and then calculating its area.

% 1. **Perimeter and Side Length**:
%    The perimeter of the equilateral triangle is given as 30 centimeters, each side is \( \frac{30}{3} = 10 \) centimeters.

% 2. **Height and Validity Check**:
%    The height (h) of an equilateral triangle can be calculated from its side length (s) using the formula for the height \( h = \frac{\sqrt{3}}{2} s \). For a side length of 10 cm, the height should be 5\sqrt{3} \approx 8.66 \text{ cm}
   
%    However, the problem states that the height is 10 centimeters. This discrepancy suggests that either the problem statement has an error or is presenting an impossible scenario since the calculated height does not match the given height.
%First, since we have an **equilateral triangle** with a perimeter of 30 centimeters, we can find the length of each side. The perimeter \(P\) of a triangle is the sum of the lengths of its sides. For an equilateral triangle, all sides are equal, so we divide the perimeter by 3 to find the length of one side.
\par...\par
% \[ P = 3 \times \text{side length} \]Given \(P = 30\) cm, we find:\[ \text{side length} = \frac{P}{3} = \frac{30}{3} = 10 \, \text{cm} \]Next, to calculate the area \(A\) of a triangle, we typically use the formula:\[ A = \frac{1}{2} \times \text{base} \times \text{height} \]Given that the height (\(h\)) is 10 cm and the base (\(b\)) is the side length we just calculated (10 cm), we substitute these values into the formula:\[ A = \frac{1}{2} \times 10 \times 10 = 50 \, \text{cm}^2 \]
However, there's a catch. The height given doesn't align with the properties of an equilateral triangle where each side is 10 cm. In an equilateral triangle, the height (\(h\)) can be found using the Pythagorean theorem for a 30-60-90 triangle, which is derived from dividing the equilateral triangle into two right triangles. The formula for the height in terms of the side length (\(a\)) is:\[ h = \frac{\sqrt{3}}{2}a \]For a side length of 10 cm, the correct height would be:\[ h = \frac{\sqrt{3}}{2} \times 10 \approx 8.66 \, \text{cm} \]So, if we were to use the correct height in calculating the area, it would actually be:\[ A = \frac{1}{2} \times 10 \times 8.66 \approx 43.3 \, \text{cm}^2 \]Therefore, with the height given as 10 cm, the problem doesn't align with the properties of an equilateral triangle with sides of 10 cm each. This discrepancy suggests a misunderstanding in the problem statement as the height and the properties of the equilateral triangle given \colorbox {green}{do not match}.
% \\
%         \hline
% Find the \colorbox {yellow}{integer} solution of the equation $x^2+x=3$. &&
\\
        \hline
        Max picks \colorbox {yellow}{five} different cards without replacement from a standard 52-card deck. What is the probability that the cards are of different suits? &
To calculate the probability that Max picks five different cards of different suits from a standard 52-card deck without replacement, we need to consider the composition of the deck and the sequence of his picks.1.
 **First Card:** %When Max picks the first card, there are no restrictions. He has 52 options out of 52, so the probability of picking a card of any suit is \( \frac{52}{52} = 1 \).2. **Second Card:** For the second card to be of a different suit than the first, Max has 39 cards that are suitable (13 cards in each of the 3 remaining suits). So, the probability for the second card is \( \frac{39}{51} \) since one card is already taken out.3. **Third Card:** Now, Max needs a card from the two remaining suits. There are 26 cards suitable for this (13 cards in each of the 2 remaining suits). The probability for the third card is \( \frac{26}{50} \).4. **Fourth Card:** For the fourth card, Max needs a card from the last remaining suit. There are 13 suitable cards left for this pick. The probability for the fourth card is \( \frac{13}{49} \).
\par...\par
5. **Fifth Card:** The fifth card can be of any suit since we're only looking for different suits in the first four cards. 
\par...
% Therefore, the probability for the fifth card is \( \frac{48}{48} = 1 \), considering the task only specifies the first four cards need to be of different suits.To find the overall probability, we multiply the probabilities of each step:
% \[ \text{Overall Probability} = \frac{52}{52} \times \frac{39}{51} \times \frac{26}{50} \times \frac{13}{49} \times 1 \]\[ \text{Overall Probability} = 1 \times \frac{39}{51} \times \frac{26}{50} \times \frac{13}{49} \]\[ \text{Overall Probability} = \frac{39 \times 26 \times 13}{51 \times 50 \times 49} \]\[ \text{Overall Probability} = \frac{13182}{124950} \]\[ \text{Overall Probability} = \frac{13182}{124950} \approx 0.1055 \]
\[ \text{Overall Probability} =...\approx 0.1055 \]
So, the probability that Max picks five different cards of different suits from a standard 52-card deck without replacement is approximately 0.1055, or \colorbox {red}{10.55\%}.
&
...\par
4. **Fourth Card:** For the fourth card, it must be one of the 13 cards left that belong to the suit not yet chosen, out of the 49 cards remaining, giving a probability of $\frac{13}{49}$.The overall probability is then calculated by multiplying these probabilities together, reflecting the sequence of choices Max makes under the condition that each card comes from a different suit:$$\text{Probability} = \frac{52}{52} \times \frac{39}{51} \times \frac{26}{50} \times \frac{13}{49}.$$However, since the original question asks about five cards from different suits in a 52-card deck, which only contains four suits, the scenario is \colorbox {green}{not achievable}.
\\
        \hline

        % \bottomrule
        \end{tabular}
        }
    %GPT-4-0125-preview对Trap problem的回答，Output without prompt 指的是在没有任何额外提示词时让模型的输出，Output with prompt指的是在提示模型题目可能无解时的回答
    \caption{Responses of GPT-4-0125-preview to Trap Problems. ``Output(w/o Notice)'' refers to the model's output when no additional prompt is provided, whereas ``Output(w/ Notice)'' denotes the outputs when the model is informed that the problem may be unreasonable. The sections highlighted in \colorbox {yellow}{yellow} delineate the distinction between original problems and trap problems. The \colorbox {green}{green} sections represent instances where the model's final answers are correct, while the \colorbox {red}{red} sections indicate where the model's final answers are incorrect.}
    \label{app_tab:output_of_ChatGPT4_trap_2}
    }
\end{table*}

\begin{table*}
    % \centering
    \setlength{\tabcolsep}{2pt}
    {
    \renewcommand\arraystretch{1.0}
    \resizebox{\linewidth}{!}{
        \begin{tabular}{|m{2cm}|m{1.5cm}|m{7cm}|m{2.5cm}|m{7cm}|}
        % \toprule
        \hline
        \makecell[c]{\textbf{Original}\\\textbf{ problem}} & \makecell[c]{\textbf{Answer}} & \makecell[c]{\textbf{ Output}}& \makecell[c]{\textbf{Conceptual}\\\textbf{ Problem}} & \makecell[c]{\textbf{ Output}}  \\
        \hline
        
In right triangle $XYZ$ with $\angle YXZ = 90^\circ$, we have $XY = 24$ and $YZ = 25$. Find  \colorbox {yellow}{$\tan Y$}.
&
$\frac{7}{24}$
&
In right triangle $XYZ$, with $\angle YXZ = 90^\circ$, we're given that $XY = 24$ and $YZ = 25$. To find $\tan Y$, we first need to identify the sides relevant to angle $Y$.In a right triangle, $\tan$ of an angle is defined as the ratio of the opposite side to the adjacent side. Here, angle $Y$ is at the vertex formed by sides $XY$ and $YZ$. 
\par...\par
% Since $\tan Y$ involves the side opposite to angle $Y$ (which is $XZ$) and the side adjacent to angle $Y$ (which is $XY$), we need to find the length of $XZ$.We can use the Pythagorean theorem to find $XZ$. The theorem states that in a right triangle, the square of the length of the hypotenuse ($YZ$) is equal to the sum of the squares of the lengths of the other two sides ($XY$ and $XZ$).Given:- $XY = 24$- $YZ = 25$The equation becomes:$$YZ^2 = XY^2 + XZ^2$$Substituting the given values:$$25^2 = 24^2 + XZ^2$$Solving for $XZ^2$:$$625 = 576 + XZ^2$$Subtract $576$ from both sides:
$$XZ^2 = 625 - 576$$$$XZ^2 = 49$$Taking the square root of both sides gives:$$XZ = 7$$Now that we have the length of $XZ$ ($7$), we can find $\tan Y$:$$\tan Y = \frac{\text{opposite to Y}}{\text{adjacent to Y}} = \frac{XZ}{XY} = \frac{7}{24}$$Therefore, $\tan Y = \colorbox {green}{$\frac{7}{24}$}$.
&
Does tan90° exist?
&
\colorbox {green}{True}.To understand why, let's break it down. The tangent function, represented as $\tan(\theta)$, is defined as the ratio of the sine of 
angle $\theta$ to the cosine of angle $\theta$. In mathematical terms, this is expressed as:$$\tan(\theta) = \frac{\sin(\theta)}{\cos(\theta)}$$When we apply this definition to $\tan(90°)$, we encounter an issue. The sine of 90 degrees is 1, but the cosine of 90 degrees is 0. Plugging these values into our formula gives us:$$\tan(90°) = \frac{\sin(90°)}{\cos(90°)} = \frac{1}{0}$$Division by zero is undefined in mathematics. Because the cosine of 90 degrees is 0, attempting to calculate $\tan(90°)$ results in an undefined operation. Therefore, it is true to say that $\tan(90°)$ does not exist due to this division by zero.
\\\hline
        An equilateral triangle has a perimeter of 30 centimeters. Calculate the area of the triangle.
        &
        $25\sqrt{3} \text{ cm}^2$
        &
        To calculate the area of an equilateral triangle, you first need to know the length of one side. Since the perimeter of the triangle is given as 30 centimeters, and an equilateral triangle has all sides of equal length, each side of the triangle is:

\[ \text{Side length} = \frac{\text{Perimeter}}{3} = \frac{30 \text{ cm}}{3} = 10 \text{ cm} \]

The formula to calculate the area \( A \) of an equilateral triangle with side length \( s \) is:

\[ A = \frac{\sqrt{3}}{4} s^2 \]

Plugging in the side length:

\[ A = \frac{\sqrt{3}}{4} \times (10 \text{ cm})^2 \]
\[ A = \frac{\sqrt{3}}{4} \times 100 \text{ cm}^2 \]
\[ A = \colorbox{green}{$25\sqrt{3} \text{ cm}^2$} \]
Let's calculate the exact numeric value of the area.

The area of the equilateral triangle is approximately \( 43.30 \) square centimeters.
    &
    Can the height of an equilateral triangle be equal to its side length?
    &
    To figure out if the height of an equilateral triangle can be equal to its side length, let's analyze the properties of an equilateral triangle.An equilateral triangle has all three sides of the same length. If we draw a height from one vertex to the opposite side, it not only acts as a height but also bisects the base into two equal parts, creating two 30-60-90 right triangles. In a 30-60-90 triangle, the sides are in the ratio of $1:\sqrt{3}:2$. Here, the smallest side (opposite the 30° angle) would be half the length of the hypotenuse (the original side of the equilateral triangle), and the height (opposite the 60° angle) would be $\sqrt{3}/2$ times the length of the hypotenuse.So, if the side length of the equilateral triangle is $s$, the height $h$ would be $s\cdot\sqrt{3}/2$. For the height to be equal to the side length, $s = s\cdot\sqrt{3}/2$, which is not true unless $s = 0$ (which is not a valid length for a side of a triangle).**Conclusion:**\colorbox {green}{False}, the height of an equilateral triangle cannot be equal to its side length.
    \\
    \hline
    Natalia sold 48 clips in April and half as many clips in May. How many clips did Natalia sell altogether in April and  \colorbox {yellow}{June}?
    &72
    &
    Alright, let's break it down:1. **April Sales**: Natalia sold clips to 48 of her friends in April.2. **May Sales**: It's mentioned that she sold half as many clips in May as she did in April. So, if she sold 48 clips in April, then in May she sold $\frac{1}{2} \times 48 = 24$ clips.To find out how many clips Natalia sold altogether in April and May, we simply add the two amounts together:$$48 + 24 = 72$$So, **Natalia sold a total of \colorbox {green}{72} clips** to her friends over the two months.
    &Knowing the sales for May and June, without knowing the sales for July, is it possible to calculate the sum of the sales for June and July?
    & \colorbox {green}{False}.To calculate the sum of the sales for June and July, you need the exact sales figures for both months. Knowing the sales for May gives you insight into the performance of that specific month but does not contribute any direct information about July's sales. Without the sales data for July, you cannot accurately determine the sum of June and July's sales. Therefore, it's not possible to calculate this sum with the information provided.
\\
        \hline
        % \bottomrule
        \end{tabular}
        }
    
    \caption{Responses of GPT-4-0125-preview to both original and conceptual problems. The sections highlighted in \colorbox {yellow}{yellow} delineate the distinction between original problems and trap problems. The \colorbox {green}{green} sections represent instances where the model's final answers are correct.}
    \label{app_tab:output_of_ChatGPT4_original_1}
    }
\end{table*}

\begin{table*}
    % \centering
    \setlength{\tabcolsep}{2pt}
    {
    \renewcommand\arraystretch{1.0}
    \resizebox{\linewidth}{!}{
        \begin{tabular}{|m{2cm}|m{1.5cm}|m{7cm}|m{2.5cm}|m{7cm}|}
        % \toprule
        \hline
        \makecell[c]{\textbf{Original}\\\textbf{ problem}} & \makecell[c]{\textbf{Answer}} & \makecell[c]{\textbf{ Output}}& \makecell[c]{\textbf{Conceptual}\\\textbf{ Problem}} & \makecell[c]{\textbf{ Output}}  \\
        \hline
        
% Find the solution of the equation $x^2+x=3$
% &
% $\frac{-1 \pm \sqrt{13}}{2}$
% &
% &
% &
% \\\hline
Max picks \colorbox {yellow}{2} different cards without replacement from a standard 52-card deck. What is the probability that the cards are of different suits?
        &
        $\frac{13}{17}$
        &
        To find the probability that Max picks two cards of different suits from a standard 52-card deck, we first need to understand the deck's composition and the process of drawing cards without replacement. 1. **Deck Composition**: A standard deck of 52 cards consists of four suits: hearts, diamonds, clubs, and spades. Each suit has 13 cards. 2. **First Card Draw**: When Max draws the first card, it can be of any suit. This event is certain, so the probability is 1. 3. **Second Card Draw**: After the first card is drawn, there are 51 cards left in the deck. Since we want the second card to be of a different suit, we need to exclude the remaining 12 cards of the same suit as the first card. This leaves us with 39 cards (51 total minus 12 of the same suit) that are of a different suit. 4. **Calculating the Probability**: The probability that the second card is of a different suit is the number of favorable outcomes divided by the total number of possible outcomes. So, the probability is \( \frac{39}{51} \). 5. **Final Probability**: Since the first card can be of any suit and does not affect the probability, the final probability that the two cards drawn are of different suits is \( 1 \times \frac{39}{51} \), which simplifies to \( \frac{39}{51} \).Therefore, the probability that Max picks two cards of different suits without replacement from a standard 52-card deck is \colorbox {green}{\( \frac{39}{51} \)}, which can be simplified further if needed.
    &
    Is it possible to pick five different suits of cards from a standard deck of playing cards?
    &
    Yes, it's possible to select five different suits of cards from a standard deck of playing cards, but there's a small catch. A standard deck of playing cards typically contains four suits: hearts, diamonds, clubs, and spades. So, if you're selecting five suits, you'd have to include all four of these plus one more. However, since there are only four suits in a standard deck, you \colorbox {green}{can't} pick five different ones. You could say you're selecting all the suits in the deck, but not five different ones.
    \\
    \hline
    
        % \bottomrule
        \end{tabular}
        }
    %Output代表GPT-4-0125-preview对于原始正常问题和概念问题直接询问的回答，其中黄色部分是正确问题和Trap problem的区别，绿色部分代表模型最终答案是正确的，红色部分代表模型最终答案是错误的
    \caption{Responses of GPT-4-0125-preview to both original and conceptual problems. The sections highlighted in \colorbox {yellow}{yellow} delineate the distinction between original problems and trap problems. The \colorbox {green}{green} sections represent instances where the model's final answers are correct.}

    \label{app_tab:output_of_ChatGPT4_original_2}
    }
\end{table*}

\end{document}